\documentclass[11pt,a4paper]{article}

\usepackage[utf8]{inputenc}
\usepackage[margin=2.5cm]{geometry}
\usepackage{microtype}
\usepackage{amsmath,amssymb,bm}
\usepackage{graphicx}
\usepackage{booktabs}
\usepackage{xcolor}
\usepackage{enumitem}
\usepackage[labelfont=bf,font=small]{caption}
\usepackage{tikz}
\usetikzlibrary{arrows.meta,positioning,calc}
\usepackage[numbers,sort&compress]{natbib}
\usepackage[colorlinks=true,linkcolor=blue!50!black,citecolor=blue!50!black,urlcolor=blue!50!black]{hyperref}
\usepackage[capitalise,noabbrev]{cleveref}
\hypersetup{pdftitle={Multiclass Boundary Refinement with Neural Cellular Automata for Remote Sensing Land Cover Segmentation},
  pdfauthor={Pascoassis Souza Santos Meira, Gilson Antonio Giraldi, Rafael Duarte Coelho dos Santos}}

\setlist{nosep,leftmargin=1.4em}
\newcommand{\nc}{n_c}
\newcommand{\Sx}{S_x}
\newcommand{\Sy}{S_y}

\definecolor{PercepBlue}{HTML}{1F4E79}
\definecolor{CondOrange}{HTML}{E07A3F}
\definecolor{RuleOrange}{HTML}{C2570F}
\definecolor{ViTInk}{HTML}{4A4E69}
\definecolor{StateInk}{HTML}{111111}
\definecolor{RGBInk}{HTML}{2E7D32}

\newcommand{\iconRGB}{%
  \begin{tikzpicture}[x=1mm,y=1mm]
    \draw[rounded corners=0.3mm, fill=blue!6, draw=black!45] (0,0) rectangle (10,8);
    \fill[yellow!85!orange] (7.7,6.2) circle (1.0);
    \fill[green!40!black] (0,0.3) -- (3.6,4.4) -- (5.6,2.6) -- (10,5.4) -- (10,0.3) -- cycle;
  \end{tikzpicture}%
}
\newcommand{\iconViT}{%
  \begin{tikzpicture}[x=1mm,y=1mm]
    \fill[black] (0,0) rectangle (10,8);
    \draw[white,line width=0.55mm] (3.5,4.3) .. controls (3.5,6.6) and (6.5,6.6) .. (6.5,4.3);
    \fill[white,rounded corners=0.3mm] (2.9,1.1) rectangle (7.1,4.5);
    \fill[black] (4.55,2.2) rectangle (5.45,3.4);
  \end{tikzpicture}%
}
\newcommand{\iconTensor}{%
  \begin{tikzpicture}[x=1mm,y=1mm]
    \foreach \i in {0,1,2}{
      \draw[fill=black!14,draw=black!45,line width=0.15mm] (1.1*\i,1.1*\i) rectangle +(6.5,6.5);
    }
    \foreach \i/\op in {3/55,4/85}{
      \draw[fill=PercepBlue!\op!white,draw=PercepBlue,line width=0.15mm] (1.1*\i,1.1*\i) rectangle +(6.5,6.5);
    }
  \end{tikzpicture}%
}
\newcommand{\miniKernel}[1]{%
  \begin{tikzpicture}[x=1mm,y=1mm]
    \draw[step=1.9mm,black!55,line width=0.12mm] (0,0) grid (5.7,5.7);
    #1
  \end{tikzpicture}%
}
\newcommand{\iconPercep}{%
  \begin{tikzpicture}[x=1mm,y=1mm]
    \node at (0,0) {\miniKernel{\fill[PercepBlue!70] (1.9,1.9) rectangle (3.8,3.8);}};
    \node at (7.2,0) {\miniKernel{\fill[PercepBlue!70] (0,1.9) rectangle (1.9,3.8); \fill[PercepBlue!25] (3.8,1.9) rectangle (5.7,3.8);}};
    \node at (14.4,0) {\miniKernel{\fill[PercepBlue!70] (1.9,3.8) rectangle (3.8,5.7); \fill[PercepBlue!25] (1.9,0) rectangle (3.8,1.9);}};
  \end{tikzpicture}%
}
\newcommand{\iconCond}{%
  \begin{tikzpicture}[x=1mm,y=1mm]
    \draw[fill=CondOrange!22,draw=CondOrange,line width=0.2mm] (0,0) -- (2.2,1.7) -- (2.2,7.7) -- (0,6) -- cycle;
    \draw[fill=CondOrange!42,draw=CondOrange,line width=0.2mm] (4.6,0) -- (6.8,1.7) -- (6.8,7.7) -- (4.6,6) -- cycle;
  \end{tikzpicture}%
}
\newcommand{\iconMLP}{%
  \begin{tikzpicture}[x=1mm,y=1mm]
    \foreach \y [count=\ii] in {1,4,7}{\coordinate (i\ii) at (0,\y);}
    \foreach \y [count=\hi] in {0,3,6,9}{\coordinate (h\hi) at (5,\y);}
    \foreach \y [count=\oi] in {2.5,6.5}{\coordinate (o\oi) at (10,\y);}
    \foreach \a in {1,...,3}{\foreach \b in {1,...,4}{\draw[black!25,line width=0.1mm] (i\a) -- (h\b);}}
    \foreach \a in {1,...,4}{\foreach \b in {1,...,2}{\draw[black!25,line width=0.1mm] (h\a) -- (o\b);}}
    \foreach \a in {1,...,3}{\fill[RuleOrange] (i\a) circle (0.55);}
    \foreach \a in {1,...,4}{\fill[RuleOrange!65] (h\a) circle (0.55);}
    \foreach \a in {1,...,2}{\fill[RuleOrange] (o\a) circle (0.55);}
  \end{tikzpicture}%
}

\title{Multiclass Boundary Refinement with Neural Cellular Automata\\
for Remote Sensing Land Cover Segmentation}

\author{Pascoassis Souza Santos Meira\textsuperscript{1} \qquad Gilson Antonio Giraldi\textsuperscript{1}\\
Rafael Duarte Coelho dos Santos\textsuperscript{2}\\[3pt]
\normalsize \textsuperscript{1}Laborat\'orio Nacional de Computa\c{c}\~ao Cient\'ifica (LNCC), Petr\'opolis, RJ, Brazil\\[3pt]
\normalsize \textsuperscript{2}Instituto Nacional de Pesquisas Espaciais (INPE), S\~ao Jos\'e dos Campos, SP, Brazil}
\date{October 2026}

\begin{document}
\maketitle

\begin{abstract}
Semantic segmentation of remote sensing imagery often places class boundaries a few
pixels away from where they truly lie, and this \emph{position error} is concentrated in the
thin band measured by Boundary IoU (BIoU). We treat boundary refinement as the iterated
application of a small learned local rule: the Boundary Refinement Neural Cellular Automaton (BRNCA), that starts from the class probabilities of a frozen, pre-trained segmenter and corrects them cell by cell.
We generalise the binary rNCA to the multiclass case by keeping the visible channels on the
probability simplex through an Aitchison (log-space) update and by adding free hidden
channels. 
The best configuration improves the mean BIoU over the base segmenter by $2.36$ points (from 46.63 to 48.99). The gain is concentrated on the irregular \emph{woodland} class ($+8.5$ points BIoU); this suggests that \emph{position error} is correlated with the fractal dimension of the class boundaries.
\end{abstract}

\section{Introduction}
\label{sec:intro}

Land use and land cover (LULC) maps underpin the monitoring of territorial change and
evidence-based environmental policy. In remote sensing, producing them is a multiclass semantic
segmentation problem: every pixel receives exactly one of several categories, so every region is
delimited by the boundaries it shares with its neighbours, and the quality of the map depends
directly on where those boundaries are placed. Modern segmenters, from convolutional
networks \cite{long2015fullyconvolutionalnetworkssemantic,ronneberger2015unet} to Vision
Transformers \cite{dosovitskiy2021vit} and foundation-model encoders
\cite{oquab2023dinov2,ranftl2021dpt}, are good at deciding \emph{what} a region is and less
reliable at deciding its boundary.

\begin{figure}[t]
  \centering
  \includegraphics[width=\linewidth]{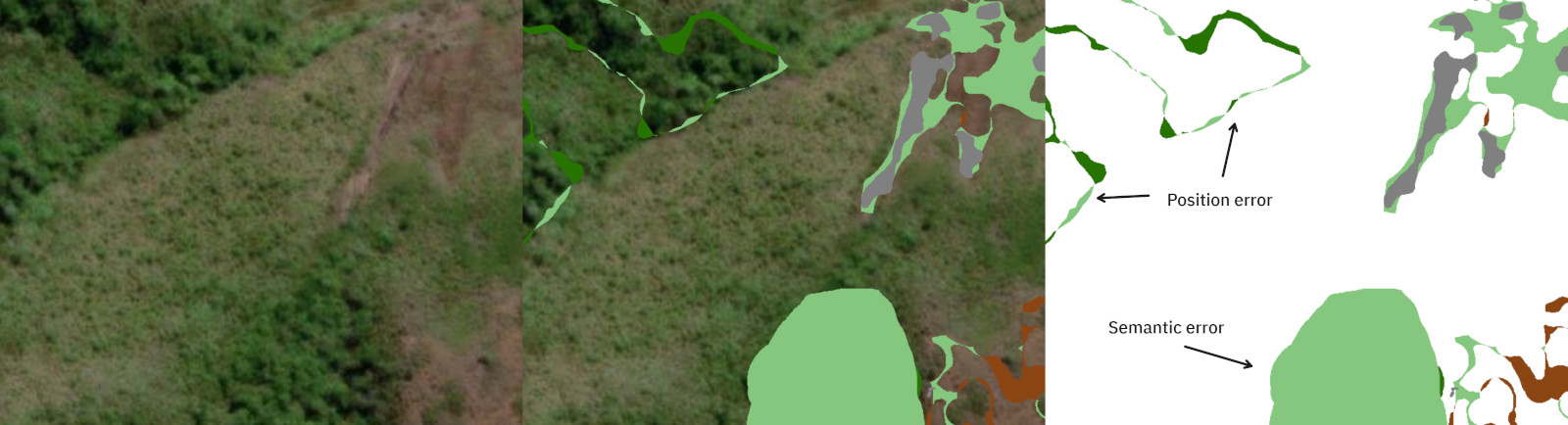}
  \caption{Position and semantic errors in a real segmentation. Left: input image; middle: error
  regions overlaid on it; right: error regions alone. Position errors are thin slivers along
  region contours; the large light-green blob is a semantic error (sparse vegetation predicted
  over an area of dense vegetation).}
  \label{fig:errors}
\end{figure}

We distinguish two kinds of error (\cref{fig:errors}). A \emph{semantic error} assigns a whole
region to the wrong class. A \emph{position error} occurs when the interiors of two neighbouring
regions are classified correctly but the decision boundary separating them is displaced with
respect to the true one. Global metrics such as IoU mix both effects and are dominated by
interiors. Boundary IoU (BIoU) \cite{cheng2021boundaryiou} restricts the computation to a
narrow band around the annotated boundary and therefore isolates position error; its
complement, Interior IoU (IIoU), measures the remainder of each region. State-of-the-art
segmenters typically score a much lower BIoU than IoU \cite{cheng2021boundaryiou}, demonstrating that boundary refinement remains an open challenge.

Two further obstacles are specific to remote sensing. First, the contours of natural classes
(vegetation, rock, water) are highly irregular and can be characterised by fractal dimension
\cite{mandelbrot1967coast,wang2024fractalcurve}. Second, a segmenter sees the scene through a finite window
(the raster crop, the backbone patch, the convolution kernel or attention window) that is much
smaller than the continuous scene it represents. We quantify the first obstacle through the fractal
dimension of the class boundaries (\cref{sec:data}) and address the second with
local, iterated refinement.

Cellular automata (CA) \cite{von1966theory,wolfram2002new} apply one simple rule uniformly to the
immediate neighbourhood of every cell. Such a rule works at the native image resolution, and can be
directed to the boundary, leaving interior decisions that are already correct untouched. A
Neural Cellular Automaton (NCA) \cite{mordvintsev2020growing} parametrises the rule with a small
neural network and trains it by back-propagation through time (BPTT)
\cite{werbos1990backpropagation}; unrolling the automaton's time steps, it can be seen as a recurrent residual convolutional network.
rNCA \cite{silbernagel2025rnca} showed that an NCA can repair binary segmentation masks. Here we generalise the idea to multiclass segmentation with the Boundary Refinement Neural Cellular Automaton (BRNCA) and ask a deliberately narrow question: \emph{what
is the smallest NCA that moves boundaries to the right place?}

\paragraph{Contributions.}
\begin{enumerate}
  \item A \textbf{multiclass boundary refinement NCA (BRNCA)} for a frozen base segmenter: the cell state
  holds the class distribution on the probability simplex, updated through an Aitchison
  (log-space) step, plus free hidden channels (\cref{sec:method}).
  \item A \textbf{minimal design} without sample pool, alive mask or stochastic updates, with the
  rationale for each removal in the setting of boundary position (\cref{sec:design}). Also, we conduct a brief categorisation of the dynamic behaviour of BRNCA.
  \item A controlled \textbf{core ablation} over the number of hidden channels
  ($m\in\{10,15,20\}$) and the number of neurons in the hidden layer of the update MLP ($d_h\in\{64,128,256\}$), with the
  original identity+Sobel perception bank, and a per-class BIoU/IIoU analysis related to the fractal
  dimension of the class boundaries (\cref{sec:ablation,sec:perclass}).
\end{enumerate}

\section{Related Work}
\label{sec:related}

\paragraph{Neural cellular automata.}
Mordvintsev et al.\ \cite{mordvintsev2020growing} introduced NCAs to grow and regenerate a
$64\times64$ RGBA image. Every cell holds a state vector that is updated, in parallel and with the
same rule, by two modules: \emph{perception}, which aggregates the neighbours with a fixed bank of
depthwise kernels (identity and Sobel), and \emph{update}, a shallow network that maps the
perception vector to a state increment. Later works prune the machinery of the original NCA: Variational NCA \cite{palm2022variational}
dispenses with the alive mask, and NoiseNCA \cite{pajouheshgar2024noisenca} additionally removes the
stochastic asynchronous update, injecting noise in the perception stage instead.

\paragraph{NCAs for segmentation.}
rNCA \cite{silbernagel2025rnca} is the closest work to our proposal. It applies an NCA to the output of a dense
segmenter to repair binary masks of medical images, conditioning the rule on the input image;
it targets topological errors (fragmented vessels, broken myocardium), measured with Dice, clDice
and Betti errors. We instead target \emph{boundary position} in the \emph{multiclass} setting of
land cover mapping.

\paragraph{Boundary refinement.}
The dense CRF \cite{krahenbuhl2011densecrf} is the classic multiclass refiner: a unary term from
the base prediction and a pairwise term that penalises label disagreement between pixels close in
space and appearance. CRF-as-RNN \cite{zheng2015crfasrnn} parametrises the CRF as a differentiable
recurrent network, and ClassWise-CRF \cite{zhu2025classwise} adapts it to remote sensing.
Model-agnostic refiners post-process the output of any segmenter: SegFix \cite{yuan2020segfix}
replaces the labels of boundary pixels with those of interior pixels, guided by predicted boundary
and direction maps, and SegRefiner \cite{wang2023segrefiner} refines coarse masks through a discrete
diffusion process.

\section{Boundary Refinement Neural Cellular Automaton (BRNCA)}
\label{sec:method}

\subsection{Setting and state}
Let $x\in\mathbb{R}^{H\times W\times3}$ be an RGB image and $g$ a segmenter with $\nc$ classes,
pre-trained and \emph{frozen}, whose softmax output $p_0=g(x)$ lies, at every pixel, on the
probability simplex $\Delta^{\nc-1}$. BRNCA is an NCA whose cells are the pixels. The state of
a cell at step $t$ is
\begin{equation}
  u_t = (p_t,\,h_t),\qquad p_t\in\Delta^{\nc-1},\quad t=0,1,\dots,T,
  \label{eq:state}
\end{equation}
where $p_t$ represents the probability classification tensor of $\nc$ \emph{visible} channels (initialised
with the base prediction) and the $m$ \emph{hidden} channels $h_t$ are initialised at zero and evolve
freely as short-term memory, without direct supervision. The refined segmentation is $p_T$.

\subsection{Update rule}
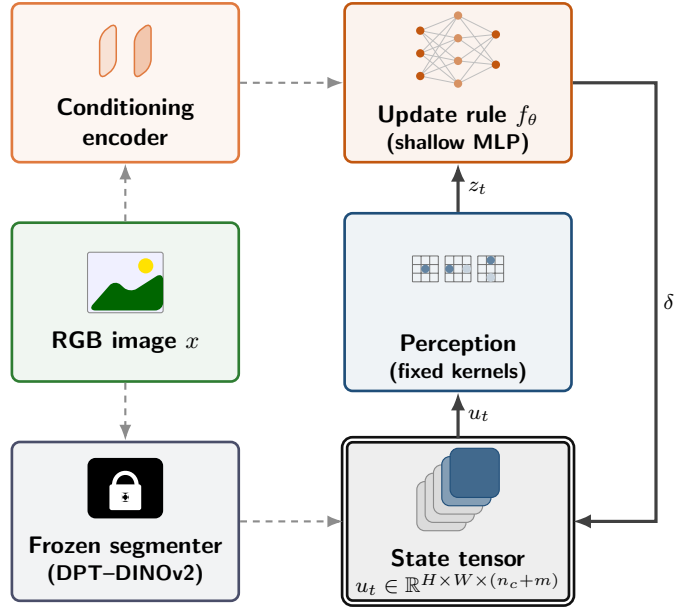
\begin{figure}[t]
\centering
\begin{tikzpicture}[
    x=1cm,y=1cm,
    >={Latex[length=2.2mm,width=1.7mm]},
    mod/.style={draw=#1,line width=1pt,fill=#1!7,rounded corners=3pt,align=center,
                font=\sffamily\bfseries\footnotesize,minimum width=3.0cm,minimum height=2.1cm,inner sep=3pt},
    lab/.style={font=\itshape\scriptsize,fill=white,inner sep=1.2pt},
    core/.style={-{Latex[length=2.4mm,width=1.9mm]},line width=1.2pt,draw=black!75},
    stat/.style={-{Latex[length=2.1mm,width=1.6mm]},line width=0.8pt,draw=black!45,dash pattern=on 2.6pt off 2.2pt}]
  \node[mod=RGBInk]        (rgb)  at (-2.4, 0)   {\iconRGB\\[2pt]RGB image $x$};
  \node[mod=CondOrange]    (cond) at (-2.4, 2.9) {\iconCond\\[2pt]Conditioning\\encoder};
  \node[mod=ViTInk]        (seg)  at (-2.4,-2.9) {\iconViT\\[2pt]Frozen segmenter\\[-1pt]\scriptsize(DPT--DINOv2)};
  \node[mod=RuleOrange]    (rule) at ( 2.0, 2.9) {\iconMLP\\[2pt]Update rule $f_\theta$\\[-1pt]\scriptsize(shallow MLP)};
  \node[mod=PercepBlue]    (per)  at ( 2.0, 0)   {\scalebox{0.6}{\iconPercep}\\[2pt]Perception\\[-1pt]\scriptsize(fixed kernels)};
  \node[mod=StateInk,double,double distance=0.9pt] (st) at ( 2.0,-2.9) {\iconTensor\\[2pt]State tensor\\[-1pt]\scriptsize$u_t\in\mathbb{R}^{H\times W\times(\nc+m)}$};
  \draw[stat] (rgb) -- (cond);
  \draw[stat] (rgb) -- (seg);
  \draw[stat] (cond) --  (rule);
  \draw[stat] (seg)  --  (st);
  \draw[core] (st)  -- node[lab,right=2pt] {$u_t$} (per);
  \draw[core] (per) -- node[lab,right=2pt] {$z_t$} (rule);
  \draw[core] (rule.east) -- ++(1.1,0) |- node[lab,pos=0.25,right=1pt] {$\delta$} (st.east);
\end{tikzpicture}
\caption{One BRNCA step. The frozen segmenter provides the initial state once; the loop (solid
arrows) perceives the state, computes an increment $\delta=(\delta^{p},\delta^{h})$ with a shallow MLP
conditioned on the image, and updates every cell synchronously (\cref{eq:update}).}
\label{fig:arch}
\end{figure}

\Cref{fig:arch} shows one step of the automaton; dashed arrows are computed once, solid arrows form
the iterated loop. Three modules act on the state.

\paragraph{Perception.}
The fixed bank $\mathcal{K}=\{\mathrm{Id},\Sx,\Sy\}$ of $3\times3$ depthwise kernels (identity and the two Sobel
kernels, as in the original NCA \cite{mordvintsev2020growing}) is applied to every channel of $u_t$, giving
the perception vector $z_t=(K*u_t)_{K\in\mathcal{K}}$.

\paragraph{Conditioning.}
A small trainable convolutional encoder maps the RGB image to a context vector $E(x)$, computed once.
It gives the rule access to appearance information that is absent from the class probabilities.

\paragraph{Update.}
A shallow MLP $f_\theta$, shared by all cells (a $1\times1$ convolution), maps $[z_t,E(x)]$ to an increment
$\delta=(\delta^{p},\delta^{h})$. Its hidden layer has $d_h$ neurons. A plain additive
step would leave the simplex, so the visible channels are updated by an Aitchison perturbation
\cite{aitchison1982statistical} in log-probability space, and the hidden channels are updated additively:
\begin{equation}
  p_{t+1}=\mathrm{softmax}\!\big(\log p_t+\delta^{p}\big),\qquad
  h_{t+1}=h_t+\delta^{h},\qquad \delta=f_\theta\big(z_t,E(x)\big).
  \label{eq:update}
\end{equation}
Therefore, the rule re-weights the current class evidence
multiplicatively, so a class can be suppressed or promoted without ever leaving the simplex.

\subsection{Design choices: a minimal BRNCA}
\label{sec:design}
Earlier NCAs for growth and repair carry extra machinery: an alive mask, a sample pool and
stochastic asynchronous updates \cite{mordvintsev2020growing,silbernagel2025rnca}. We drop all three components.
These are design decisions justified by our setting; they are not part of the core ablation of \cref{sec:ablation}.

\paragraph{No alive mask.}
In Growing NCA the alive mask separates cells that take part in growth from empty background, and rNCA
reuses its visible probability channel for that role. Variational NCA \cite{palm2022variational} and
NoiseNCA \cite{pajouheshgar2024noisenca} dispense with it. In our setting the initial state is the dense
output of a segmenter, not a sparse seed, so every cell carries a meaningful class distribution from
$t=0$ and there is no growth front to gate. There is a second argument specific to boundary position.
When class boundaries are misaligned, the incorrectly included pixels typically belong to regions classified with high confidence, meaning the model is sure of the category but blind to its precise limits. A mask driven by the state, whether by confidence or by margin, would tend to freeze exactly the cells that
must change class for the boundary to move, and would tie the region where the boundary may move to the
region where the base model already placed it. Leaving every cell free lets $f_\theta$ decide, from
perception and conditioning, where to act.

\paragraph{No sample pool.}
The pool maintains previously evolved state tensors to stabilise training across long trajectories and promote pattern persistence, feeding back a mixed batch comprising states sampled from the pool alongside fresh predictions generated directly by the base segmenter.
However, in our boundary refinement setting, every training iteration starts directly from the base segmenter's output $p_0$ and a zero hidden state, matching the inference procedure while eliminating additional RAM--VRAM transfer overhead.
Without the pool, the rule is never trained on states beyond the maximum training step ($T=64$); we measure the impact of this choice in \cref{sec:beyondT}.

\paragraph{Synchronous updates.}
Unlike Growing NCA, which updates cells asynchronously, governed by a stochastic update probability and gated by an alive mask, in BRNCA every cell updates synchronously at every time step ($p_{\text{stoch}} = 1$) without pixelwise dropout or spatial gating.

\subsection{Training and inference}
The automaton runs for $T$ steps and is trained by BPTT with the cross-entropy between the final visible
channels $p_T$ and the ground truth, averaged over pixels (computed as the negative log-likelihood of $p_T$, which already lies on the simplex). During training, the number of automaton steps $T$ is set between $32$ and $64$ (so $T=64$ is the maximum training step); at inference, $T=64$. The base segmenter stays frozen throughout.

\section{Experiments}
\label{sec:experiments}

\subsection{LandCover.ai Dataset}
\label{sec:data}
\begin{figure}[t]
    \centering
    \includegraphics[width=0.65\linewidth]{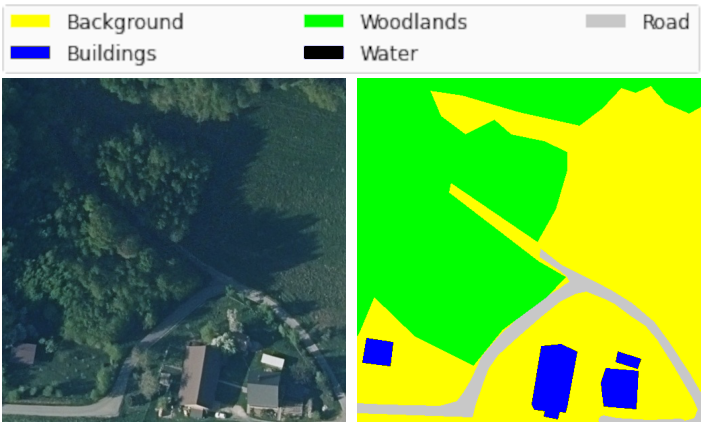}
    \caption{Cropped example of a LandCover.ai aerial image and its annotation mask, with colour legend for the five segmentation classes.}
    \label{fig:landcoverai}
\end{figure}

LandCover.ai \cite{boguszewski2021landcoverai} consists of aerial orthophotos of Poland (41 images, 25--50\,cm/pixel; 8 images of $4200\times4700$ and 33 of $9000\times9500$ pixels) with five classes: background, building,
woodland, water and road. It is the benchmark on which the core ablation is performed in \cref{sec:ablation}. \Cref{fig:landcoverai} illustrates the pixelwise annotation.

\paragraph{Boundary complexity.}
To quantify how intricate the annotated boundaries are, we estimate the fractal dimension $D$ of each class boundary
by box counting on the contour of the binary class-versus-rest mask (extracted with a morphological gradient),
fitting $\log N(r)=D\log(1/r)+C$ by least squares, where $N(r)$ is the number of boxes of side $r$ needed to cover
the contour. For the LandCover.ai dataset, the per-class values range from $D=1.13$ (water) to $D=1.33$ (background), so the classes differ
moderately, but enough to be ordered: water $<$ building $<$ road $<$ woodland $<$ background.

Boundary complexity is computed on the annotated masks at their native resolution, serving as a key factor for appropriate boundary refinement. The high $D$ value for the background class occurs because it represents the complement of the other four classes; consequently, its contour consists of the interfaces with all of them, meaning $D$ does not reflect the property of a natural land-cover class.

\subsection{Setup}
The base segmenter is a Dense Prediction Transformer \cite{ranftl2021dpt} with a DINOv2 encoder \cite{oquab2023dinov2} (DPT--DINOv2), kept frozen. It operates with a input size of $518 \times 518$, whereas BRNCA processes four $259 \times 259$ sub-windows that are subsequently reconstructed for direct comparison. On LandCover.ai, the base segmenter achieves 83.69\% mIoU and 46.63\% BmIoU (\cref{tab:ablation}), establishing the floor that BRNCA must improve. We report per-class IoU, BIoU and IIoU along with their class averages (mIoU, BmIoU, mIIoU); gains $\Delta$ represent differences relative to the base segmenter, expressed in percentage points (pp). BIoU is computed on a band of width $d=5$ pixels around the annotated contour. We use
AdamW \cite{loshchilov2019adamw} with learning rate $10^{-4}$. Reported numbers correspond to the epoch with the best BmIoU, with early stopping (patience of 10 epochs) and a maximum of 60 epochs. Experiments ran on single nodes of the SDumont supercomputer (LNCC), each with 6 CPU cores per task, 64\,GB of RAM, and four 32\,GB Tesla V100 GPUs.

\subsection{Core ablation}
\label{sec:ablation}
The minimal BRNCA of \cref{sec:design} has two free structural choices, which define its core: the number of
\emph{hidden channels} $m$ (memory carried by each cell) and the number of neurons $d_h$ in the hidden layer of the update MLP (capacity of the rule). The perception bank is fixed to identity+Sobel. We evaluate all pairs $(m,d_h)$ with
\[
m\in\{10,15,20\},\qquad d_h\in\{64,128,256\},
\]
that is, nine configurations, with everything else fixed (synchronous updates, no alive mask, no pool, optimiser and number of steps as in \cref{sec:method}). The experiment addresses two questions: (i) whether every core configuration improves BIoU over the base segmenter; and (ii) how sensitive the gain is to memory ($m$) and to the capacity of the MLP ($d_h$).
\begin{table}[!ht]
\centering\small
\caption{Core ablation on LandCover.ai: mIoU (left) and BmIoU (right), in \%, for $m$
hidden channels in the cell state and $d_h$ neurons in the hidden layer of the update MLP. Bold: best cell of each metric.}
\label{tab:ablation}
\begin{minipage}[b]{0.48\textwidth}
\centering
\textbf{mIoU base segmenter = 83.69}\\[0.5em]
\begin{tabular}{lccc}
\toprule
\textbf{mIoU} & $d_h=64$ & $d_h=128$ & $d_h=256$ \\
\midrule
$m$ = 10 & 84.04 & 83.98 & 84.23 \\
$m$ = 15 & 83.86 & 84.08 & 84.23 \\
$m$ = 20 & 84.05 & 84.23 & \textbf{84.31} \\
\bottomrule
\end{tabular}
\end{minipage}
\hfill
\begin{minipage}[b]{0.48\textwidth}
\centering
\textbf{BmIoU base segmenter = 46.63}\\[0.5em]
\begin{tabular}{lccc}
\toprule
\textbf{BmIoU} & $d_h=64$ & $d_h=128$ & $d_h=256$ \\
\midrule
$m$ = 10 & 48.41 & 48.11 & 48.70 \\
$m$ = 15 & 47.56 & 48.43 & \textbf{48.99} \\
$m$ = 20 & 48.33 & 48.82 & 48.89 \\
\bottomrule
\end{tabular}
\end{minipage}
\end{table}

According to \Cref{tab:ablation}, every one of the nine configurations improves on the base segmenter in both metrics.
BmIoU rises by 0.93 to 2.36\,pp (from 46.63 to between 47.56 and 48.99), while mIoU changes by $+0.17$
to $+0.62$\,pp (from 83.69 to between 83.86 and 84.31). The gain is thus about four times larger on
the boundary band than on the whole region, as expected from BRNCA, which moves contours and leaves the semantics of the interiors to the base model. The best BmIoU (48.99) is obtained with $m=15$, $d_h=256$ and the best
mIoU (84.31) with $m=20$, $d_h=256$; the per-class analysis below uses the former. In absolute terms the position error
remains large.

\paragraph{Sensitivity to $m$ and $d_h$.}
From \Cref{tab:ablation}, averaged over $m$, BmIoU increases monotonically with $d_h$ (48.10, 48.45 and 48.86 for $d_h=64, 128, 256$), while $m$ has a weaker, non-monotonic effect (48.41, 48.33 and 48.68 for $m=10, 15, 20$). This indicates that contour refinement is limited more by local rule capacity than by auxiliary memory. Even the smallest configuration ($m=10$, $d_h=64$) captures 75\% of the maximum gain. Overall, the ablation robustly shows positive gains across all cells, driven mainly by $d_h$. This shows that performance depends more on the number of neurons in the hidden layer ($d_h$) than on the number of hidden channels ($m$).

\subsection{Where does the gain come from?}
\label{sec:perclass}

\begin{table}[!ht]
\centering\small
\caption{Per-class change with respect to the base segmenter, LandCover.ai validation, best epoch, for the
best configuration of the ablation by BmIoU ($m=15$, $d_h=256$; \cref{tab:ablation}), in pp. $D$ is the fractal dimension of
the class boundary (\cref{sec:data}). Classes are sorted by increasing $D$; the last row is the unweighted mean over
classes ($\Delta$mIoU, $\Delta$BmIoU, $\Delta$mIIoU).}
\label{tab:perclass}

\centering
\begin{tabular}{lcccc}
\toprule
\textbf{Class} & $D$ & $\Delta$IoU & $\Delta$BIoU & $\Delta$IIoU\\
\midrule
water      & 1.129 & $+0.01$ & $+0.21$ & $\phantom{-}0.00$\\
building   & 1.192 & $+0.04$ & $+0.39$ & $-0.02$\\
road       & 1.260 & $-0.05$ & $+0.82$ & $-0.04$\\
woodland   & 1.264 & $+2.15$ & $\mathbf{+8.49}$ & $-0.79$\\
background & 1.330 & $+0.55$ & $+1.89$ & $-0.32$\\
\midrule
\textbf{Mean} & 1.235 & $+0.54$ & $+2.36$ & $-0.23$\\
\bottomrule
\end{tabular}
\end{table}

\Cref{tab:perclass} decomposes the best configuration by class and relates the change in BIoU to the fractal dimension $D$
of each class boundary. The gain is far from uniform. \emph{Woodland} alone gains 8.49\,pp of BIoU; \emph{background} follows with 1.89\,pp, then \emph{road} (0.82),
\emph{building} (0.39) and \emph{water} (0.21). Without woodland the mean gain falls from 2.36 to 0.83\,pp, so the
headline BmIoU is largely coming from the natural irregular class woodland. Global IoU follows the same ordering at a smaller scale (woodland $+2.15$,
background $+0.55$; building, water and road within $\pm0.05$) and does not decrease appreciably for any class.

\paragraph{Relation to the fractal dimension.}
The classes are ordered by BIoU gain almost as they are ordered by $D$: water, building and road come first in the
same order in both rankings, and the two classes with the highest $D$ come last, with woodland ($D=1.264$) gaining more than
background ($D=1.330$). We notice that the Spearman coefficient is $\rho=0.90$, and $\rho=1$ if woodland is left out. However, the Spearman coefficient is merely ordinal, in contrast with the magnitude captured by the Pearson coefficient $r=0.38$. So $D$ is not directly correlated to how large the woodland BIoU gain is, since road and woodland have almost the same dimension (1.260 against 1.264), but gains that differ by a factor of ten (0.82 against 8.49). We conjecture that the pixels that fall inside the BIoU band help explain this discrepancy. 

Road is thin and tubular, meaning a large share of its area lies within the band, keeping the band metric close to the global one; furthermore, its remaining contour is regular, which the base segmenter already localises well. By contrast, woodland forms massive, grouped patches, so only a small share of its pixels resides in the band, concentrating the base model's errors there. 
Background is a special case, since an
improvement of the contours of building, road or woodland is also an improvement of the contour of the background. 

\paragraph{Boundary gain and interior degradation.}
The interior is not free. IIoU does not increase for any class and drops most where the boundary gain is largest: woodland
$-0.79$, background $-0.32$, road $-0.04$, building $-0.02$, water $0.00$. The ranking by interior loss is exactly the ranking
by BIoU gain (Spearman $\rho=-1$ between $\Delta$BIoU and $\Delta$IIoU, Pearson $r=-0.98$): the classes that benefit most from
repositioning the contour are those whose interiors lose most. The loss is small against the gain (on average $-0.23$\,pp against
$+2.36$\,pp) and BIoU and IIoU are measured on different pixel populations, so they cannot simply be subtracted; the
commensurable evidence is that the global mIoU rises ($+0.54$\,pp), so the net effect over all pixels is positive. One plausible reading is that a rule shared by all cells, when it moves a contour, also perturbs
interior cells, producing isolated misclassified pixels or small spurious regions within previously correct areas.

\subsection{How the automaton evolves}
\label{sec:qual}

\begin{figure}[p]
\centering
\includegraphics[width=\linewidth]{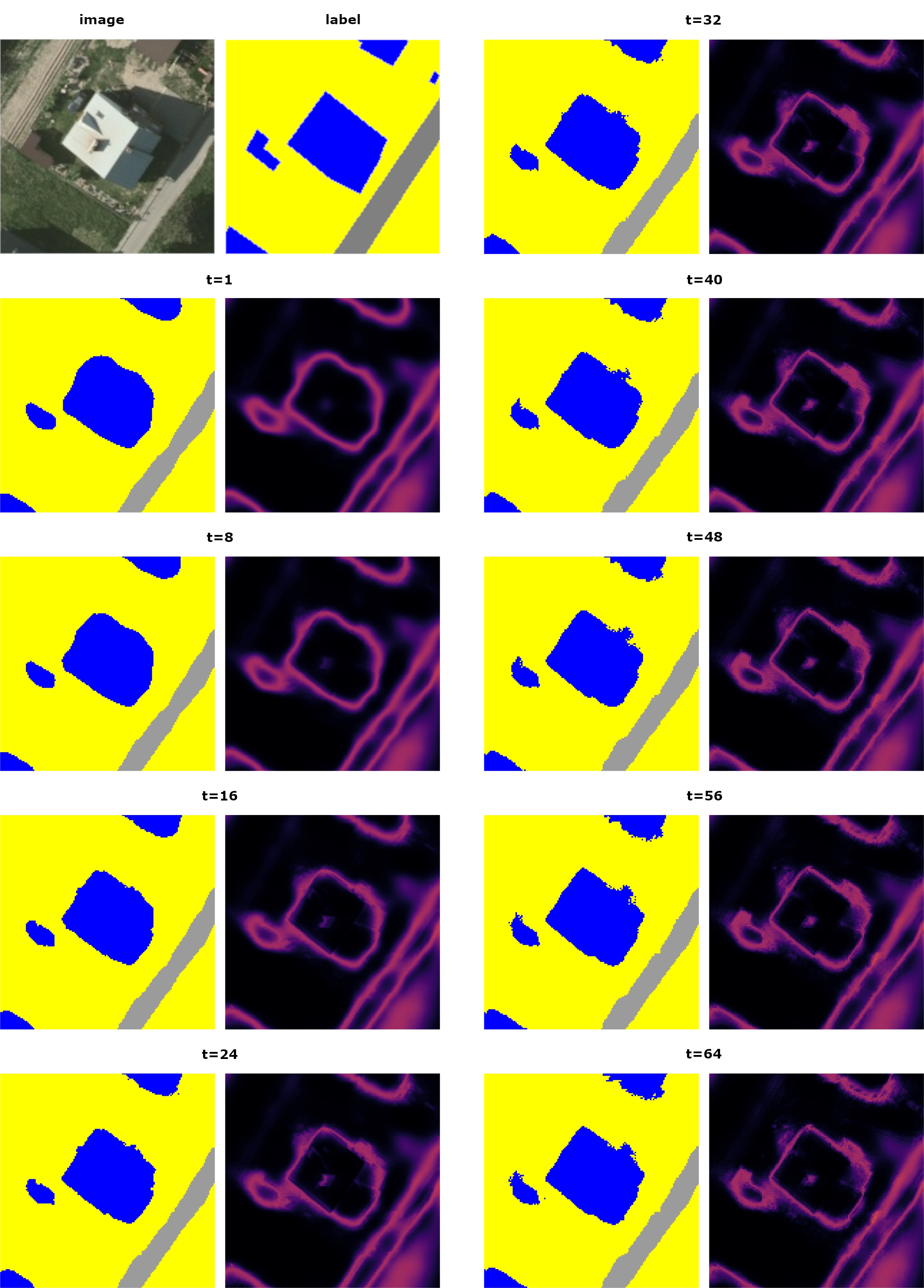}
\caption{Evolution of the automaton over its 64 steps on LandCover.ai.
Panels are read column by column, left column first. Top left: input image and ground-truth label (yellow: background; blue: building; grey: road). The
remaining panels show, for $t\in\{1,8,16,24,32,40,48,56,64\}$, the predicted label (left) and the uncertainty map (right;
dark: low, bright: high; defined by the entropy of $p_t$). }
\label{fig:evolution}
\end{figure}

\Cref{fig:evolution} follows one crop through the 64 steps used at inference. The crop focuses on anthropogenic classes (buildings, a road and the background around them). After one step ($t=1$, when the gain over the base is still only 0.03\,pp, so
the panel is essentially the base prediction) the building is a smooth, rounded blob with a wide, diffuse halo of
uncertainty. The largest changes occur between $t=8$ and $t=32$: the
building acquires straight sides and sharper corners, closer to the polygonal footprint of the label, the edge of the road
straightens, and the uncertainty collapses from a broad halo into a narrow line that follows the contour. After $t=32$
the changes are subtle at this scale, although the BmIoU gain keeps growing until about $t=62$ (\cref{fig:beyondT}), so the
later steps refine details that are hard to see in a single crop. The figure also shows some drawbacks. Along the north-east side of the building the uncertainty stays broader and the contour ragged, and from about $t=40$
small groups of isolated pixels appear near some contours (top of the crop, small blob at the left of the building).

\subsection{Behaviour beyond the maximum training step}
\label{sec:beyondT}

\begin{figure}[t]
\centering
\includegraphics[width=0.85\linewidth]{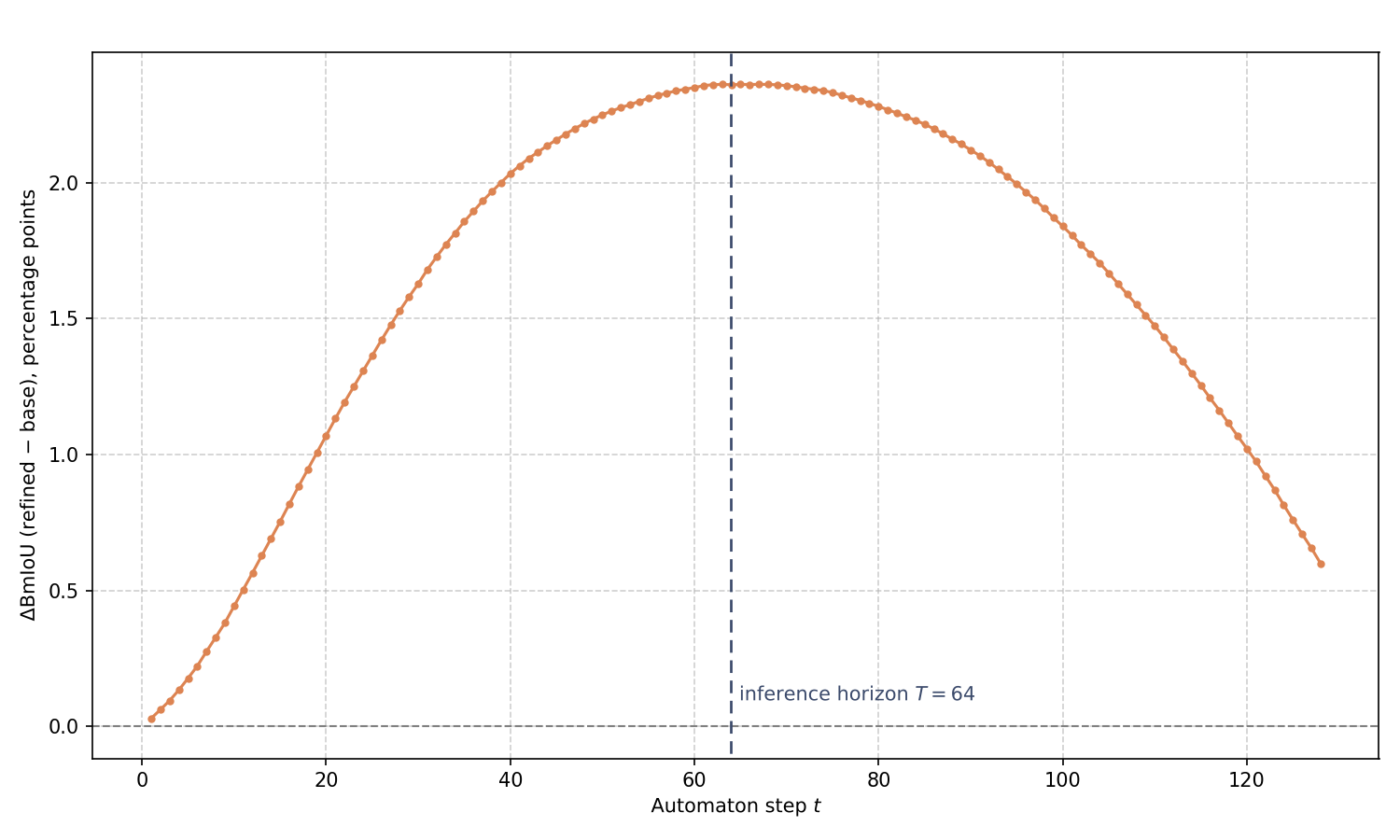}
\caption{Gain of mean BIoU over the base segmenter as a function of the number of automaton steps, for the best
configuration ($m=15$, $d_h=256$), run for twice the 64 steps used at inference. The gain peaks
at 2.36\,pp, in the plateau between roughly $t=60$ and $t=72$, and falls to about 0.6\,pp at $t=128$.}
\label{fig:beyondT}
\end{figure}

The automaton is trained and evaluated for a fixed number of steps ($T=64$ at inference). \Cref{fig:beyondT} runs the best configuration for twice that number of steps and plots the BmIoU gain at
every step. The gain grows monotonically up to a plateau between roughly $t=60$ and $t=72$, where it reaches the 2.36\,pp
of \cref{tab:ablation}, and then declines: about 1.96\,pp at $t=96$ and about 0.60\,pp at $t=128$, a quarter of the peak, still
positive but heading toward deterioration. Three things follow. First, the maximum training step of $T=64$, fixed beforehand, falls within the plateau that is best a posteriori. Second, the dynamics is not
a convergence to a fixed point that preserves the refined contour; the automaton behaves as a transient that is good at the time
it was trained for and that keeps changing, in the wrong direction, afterwards. 
Third, the decay is gradual compared to the rise in \cref{fig:beyondT}, meaning BRNCA does not diverge abruptly; nevertheless, running inference past the maximum training step ($T=64$) departs from the training objective and results in a performance loss.

This is the price of the minimal design. Without the sample pool the rule is never trained on states beyond the maximum training step, so those it meets at $t>64$ are outside the training distribution and its behaviour is unconstrained. The same effect
motivated the pool in Growing NCA, where it was introduced to make the pattern persist \cite{mordvintsev2020growing}.

\section{Discussion}
\label{sec:discussion}

\paragraph{What the evidence supports.}
A small, minimal BRNCA applied on top of a frozen segmenter raises BmIoU on LandCover.ai by 0.9 to 2.4\,pp while mIoU
changes by 0.2 to 0.6\,pp, in all nine configurations of the ablation. The effect is therefore robust to the choice of $m$ and
$d_h$. However, the structural choice that clearly matters is the capacity of
the rule. The gain is selective: it is concentrated on woodland and is small (below 1\,pp) for building, water and road, and the
ranking of the classes by gain broadly follows the ranking by the fractal dimension of their boundaries. Because the interior metric falls in the classes where the boundary metric rises, BRNCA behaves as a re-allocation of accuracy toward the boundary band rather than a free
improvement, although the net effect over all pixels is positive. The scale of the effect is modest: BmIoU stays below 50\%, and BRNCA removes only a small part of the position error of the base segmenter.

\paragraph{What the dynamics tells us about the design.}
The qualitative view and the behaviour past $T=64$ shed light on two key aspects of the design choices in \cref{sec:design}. First, since every cell updates without an alive mask or pixelwise dropout, the rule is free to shift wrong contours, which is precisely what boundary correction requires; the same choice generates an interior perturbation, causing IIoU degradation by modifying non-boundary regions. Second, dropping the sample pool makes training computationally efficient, and the automaton still achieves its peak performance at the maximum training step ($T=64$), but it leaves the dynamics beyond that step uncontrolled (\cref{sec:beyondT}).

\paragraph{Scope and limitations.}
This study relies on a single frozen base segmenter (DPT--DINOv2) evaluated on one benchmark (LandCover.ai), using three values per axis ($m$ and $d_h$) in the core ablation. The conditioning encoder and the number of automaton steps are held fixed.
The alive mask, the sample pool and the asynchronous update are removed by design and not ablated, so their effect on boundary
position is argued but not measured here, apart from the indirect evidence of \cref{sec:beyondT}. The per-class conclusions rest on
five classes and a single configuration, and the explanations offered for them (the share of pixels inside the band, small-scale
changes along fractal contours, perturbation of interior cells) are conjectures that this study did not test. BIoU is computed for
a single band width. Finally, the mean gains are of the order of one to two BIoU points, and differences of a few tenths of a point
between cells are not interpreted. Comparisons with other boundary refiners and the computational cost of the refinement are not
reported in this version.

\paragraph{Future work.}
Three directions follow. First, restoring the sample pool and studying the convergence and the attractors of the learned dynamics,
so as to obtain a stopping criterion more robust than the present number of steps ($T=64$) and to let BRNCA refine the boundaries beyond this value. Second, a comparison with the boundary refiners reviewed in \cref{sec:related} (for instance dense CRF, SegFix and
SegRefiner) on at least two base segmenters, a convolutional U-Net with an EfficientNet-B7 encoder and DPT--DINOv2, reporting
$\Delta$IoU, $\Delta$BIoU and computational cost. Third, evaluation on further land-cover data, including a new dataset of the
APA Petr\'opolis protected area built in partnership with ICMBio \cite{icmbio_apa_petropolis}, which has 8 classes without a \textit{background}, and high annotation precision in boundary position.

\section{Conclusion}
We introduced BRNCA, which casts multiclass boundary refinement as the iteration of a small learned local rule on top of a frozen segmenter and reduces the NCA to its core: class probabilities kept on the simplex by an Aitchison update, free hidden channels, a fixed
perception bank and synchronous updates, with no sample pool or alive mask. On LandCover.ai all nine configurations of the $m\times d_h$ ablation improve BmIoU over a DPT--DINOv2 base, by 0.9 to 2.4\,pp (46.63 to 48.99 for the best, $m=15$, $d_h=256$),
with mIoU changing by $+0.2$ to $+0.6$\,pp. The capacity of the update rule matters more than the number of hidden channels, and
the smallest configuration already keeps three quarters of the best BIoU gain. The gain is concentrated on woodland ($+8.5$\,pp
BIoU, 72\% of the total over classes); the classes are ordered by gain almost as they are ordered by the fractal dimension of
their boundaries, and the gain is accompanied by a small loss of interior IoU that is larger where the boundary gain is larger.
The gain peaks at the number of steps used in training ($T=64$ in this case) and decays when the automaton runs longer.

\section*{Acknowledgements}
We thank the National Laboratory for
Scientific Computing (LNCC) for access to the SDumont supercomputer, and the agencies CAPES and FACC for financial support.

\bibliographystyle{plainnat}
\bibliography{refs}

\end{document}